# Dynamic Snake Upsampling Operater and Boundary-Skeleton Weighted Loss for Tubular Structure Segmentation

Yiqi Chen, Ganghai Huang*, Sheng Zhang, Jianglin Dai

Yiqi Chen ORCID: 0009-0007-8904-0381
Ganghai Huang ORCID: 0000-0001-5235-1986

*Abstract*—Accurate segmentation of tubular topological structures (e.g., fissures and vasculature) is critical in various fields to guarantee dependable downstream quantitative analysis and modeling. However, in dense prediction tasks such as semantic segmentation and super-resolution, conventional upsampling operators cannot accommodate the slenderness of tubular structures and the curvature of morphology. This paper introduces a dynamic snake upsampling operators and a boundary-skeleton weighted loss tailored for topological tubular structures. Specifically, we design a snake upsampling operators based on an adaptive sampling domain, which dynamically adjusts the sampling stride according to the feature map and selects a set of subpixel sampling points along the serpentine path, enabling more accurate subpixel-level feature recovery for tubular structures. Meanwhile, we propose a skeleton-to-boundary increasing weighted loss that trades off main body and boundary weight allocation based on mask class ratio and distance map, preserving main body overlap while enhancing focus on target topological continuity and boundary alignment precision. Experiments across various domain datasets and backbone networks show that this plug-and-play dynamic snake upsampling operator and boundary-skeleton weighted loss boost both pixel-wise segmentation accuracy and topological consistency of results. The code is available at https://github.com/YiqiChen02/DSU

## I. INTRODUCTION

FEATURE upsampling for topological tubular structures constitutes a critical step in dense prediction tasks, bearing significant importance for enhancing the model's capability to restore detailed tubular structure features and the completeness of geometric representation. This component plays a crucial role in numerous practical applications: In medical imaging analysis [1,2], high-precision vascular pixel-level reconstruction relies on effective upsampling strategies to support hemodynamic simulations and submillimeter surgical navigation; In the fields of engineering and disaster early warning [3,4], structure-aware upsampling facilitates precise segmentation of fracture networks, enabling quantitative characterization of vulnerable zones. In remote sensing applications [5-7], refined image segmentation tasks similarly depend on high-quality feature upsampling to meet the spatial accuracy requirements of urban planning and land resource assessment.

Historically, scholars have proposed solutions from multiple dimensions for subpixel-level reconstruction of complex structures. These approaches primarily include: (1) Model architecture design: Principally including feature extraction [8-13], feature fusion [14,15], and holistic framework paradigm design [16]; (2) Upsampling operator design: Principally including traditional interpolation and learnable operators [17,18], dynamic filters [19-22], frequency-domain and implicit representation methods [23,24], and generative enhancement[25]; (3) Loss strategies: Principally including basic pixel-level loss[26,27], structure-sensitive loss[28-31], and adaptive weighted loss[32].

Unlike dense prediction tasks for conventional structures, tubular structure analysis extends traditional "pixel-level prediction" to "topological consistency reconstruction". The core challenge of such tasks lies in the requirement that the upsampling process must not only restore local geometric details but also maintain topological consistency, including connectivity of topological structures and the alignment accuracy of structural boundaries. Specifically:

(1) Fine-grained feature preservation dilemma: Subpixel-level features of tubular structures (e.g., contours and edge textures) in low-resolution images are highly susceptible to information annihilation during downsampling, resulting in insufficient local semantic cues for upsampling.

(2) Cross-scale topological consistency challenge: Multi-scale upsampling typically relies on progressive reconstruction

---

- Yiqi Chen, Ganghai Huang* (corresponding author), Sheng Zhang, and Jianglin Dai are with the School of Civil Engineering, Central South University, Changsha, Hunan 410075, China.

- E-mail: huangganghai@csu.edu.cn;
Tel: +86-189-3240-6644.
- This study was financially supported by the National Natural Science Foundation of China (Nos. 52178377).

through local receptive fields, making it difficult to model cross-scale topological dependencies, which often leads to branch discontinuities or aberrant connections.

(3) Insufficient boundary alignment accuracy: Excessive focus on regional overlap metrics at the expense of edge gradient optimization causes edge distortion or misalignment.

These issues stem from the inherent mismatch between the relatively fixed kernel coverage areas of conventional upsampling operators and the long-range curvature characteristics of tubular structures, compounded by class imbalance problems that force models to suppress gradient updates in disadvantaged regions to improve overall IoU metrics. Furthermore, the insufficient topological sensitivity of standard loss functions exacerbates risks of edge blurring and structural discontinuities, ultimately creating a tripartite optimization challenge involving regional overlap, structural continuity, and boundary alignment precision. This demands upsampling operators capable of: Effectively addressing feature information loss caused by limited receptive fields, while avoiding computational explosion and attention dispersion from excessive receptive field expansion. Such requirements impose stricter demands on the feature sensing and screening ability and the structured feature retention mechanism of the operator. At the same time, it is also required that the model can pay balanced attention to the boundary, the skeleton and the main structural part in between.

To address these challenges, we propose a Dynamic Snake Upsampling Operator(DSU) and a Boundary-Skeleton Weighted Loss(BSWL).

(1) To capture the long-range curvature features of tubular structures, we propose a dynamic snake upsampling operator based on the concept of imposing continuity constraints derived from dynamic snake convolution. This upsampling operator imposes continuity constraints on sampling kernels and enhances geometric structure perception through morphological guidance. Compared to other dynamic upsampling operators, it demonstrates superior adaptability to geometric deformation in subpixel-level reconstruction tasks.

(2) To enhance multi-scale cross-level feature reconstruction capability, this study designs a differentiable dynamic adjustment module that adaptively optimizes stride lengths based on feature maps, improving the model's adaptability to diverse feature scales. To ensure symmetry in the sampling point set, the module incorporates odd-number mapping while maintaining gradient flow during backpropagation via STE (Straight-Through Estimator). This module establishes a co-constraint mechanism with snake sampling: While snake sampling maintains spatial continuity of the sampling kernels, the dynamic stride adjustment module adaptively modifies the kernel's extension range based on feature scales.

(3) To address the co-optimization challenge of regional overlap, edge alignment accuracy, and structural continuity caused by class imbalance, this study proposes a precomputed BSWL. For mask target regions, this strategy progressively assigns weights from the skeleton to boundaries based on distance fields and generates corresponding maps. Compared to conventional boundary- or skeleton-sensitive strategies, the proposed approach is: 1) Precomputed rather than network-generated, 2) Continuous rather than discrete.

With nearly negligible precomputation overhead, it simultaneously ensures balanced attention to boundary regions, skeletal structures, and intermediate transitional areas. This effectively enhances the model's tubular feature reconstruction capability while maintaining compatibility with other structure-sensitive loss functions for cooperative optimization.

This study conducts comprehensive evaluations on semantic segmentation benchmark datasets (DeepCrack[33] and DRIVE[34]) using diverse backbone architectures (UNet[14], TransUNet[35], and DSCNet[9]). The multi-metric assessment verifies the universal effectiveness of both the dynamic snake upsampler (DSU) and boundary-skeleton weighted loss (BSWL), demonstrating their significant application potential in medical image analysis and engineering structural health monitoring.

## II. RELATED WORK

### 2.1 Feature Upsampling

Upsampling operators for feature maps have been developed to meet the need for spatial resolution restoration within neural networks. Their core function is to support the progressive reconstruction of feature resolution in dense prediction models. To overcome the limitations of traditional upsampling methods [18,36] that rely on fixed interpolation rules, researchers have introduced learnable upsampling techniques. For example, transposed convolution [17] performs deconvolution with learnable parameters; however, it tends to introduce checkerboard artifacts, which distort feature distributions and interfere with semantic information.

To further explore, dynamic upsampling methods enhance adaptability for visual tasks through content-aware mechanisms. For instance, CARAFE [19] employs a subnet to generate position-adaptive dynamic convolution kernels, while FADE[20] further optimizes dynamic kernel generation through multi-scale feature fusion. Dysample [21] adaptively shifts sampling points using contextual information, granting the sampling kernel greater flexibility. Although these methods break the fixed grid-based rules of traditional upsampling via dynamic kernels, their sampling ranges remain relatively fixed-shape neighborhoods, making them inadequate for adapting to complex structures and performing effective information selection. These limitations become particularly pronounced when reconstructing topological tubular structures (e.g., blood vessels, cracks), where the square (or quasi-square[21]) sampling windows of existing methods show significant constraints.

To establish long-range feature dependencies, SAPA [22] generates perception kernels by computing feature similarity within customized windows, thereby building remote feature associations. Upsampling operators based on implicit representations [23] leverage continuous signal modeling to capture global contextual information. Numerous studies have also employed attention mechanisms [37] and generative approaches [25] to construct upsampling operators, demonstrating superior performance in feature fusion and key information capture. However, these methods typically incur substantial computational overhead and introduce excessive redundant features, leading to attention distribution dilution. To resolve the conflict between long-range dependency modeling and attention focusing, this study proposes an upsampling operator with adaptive sampling ranges (for long-range dependency features) guided by morphological constraints (for attention concentration).

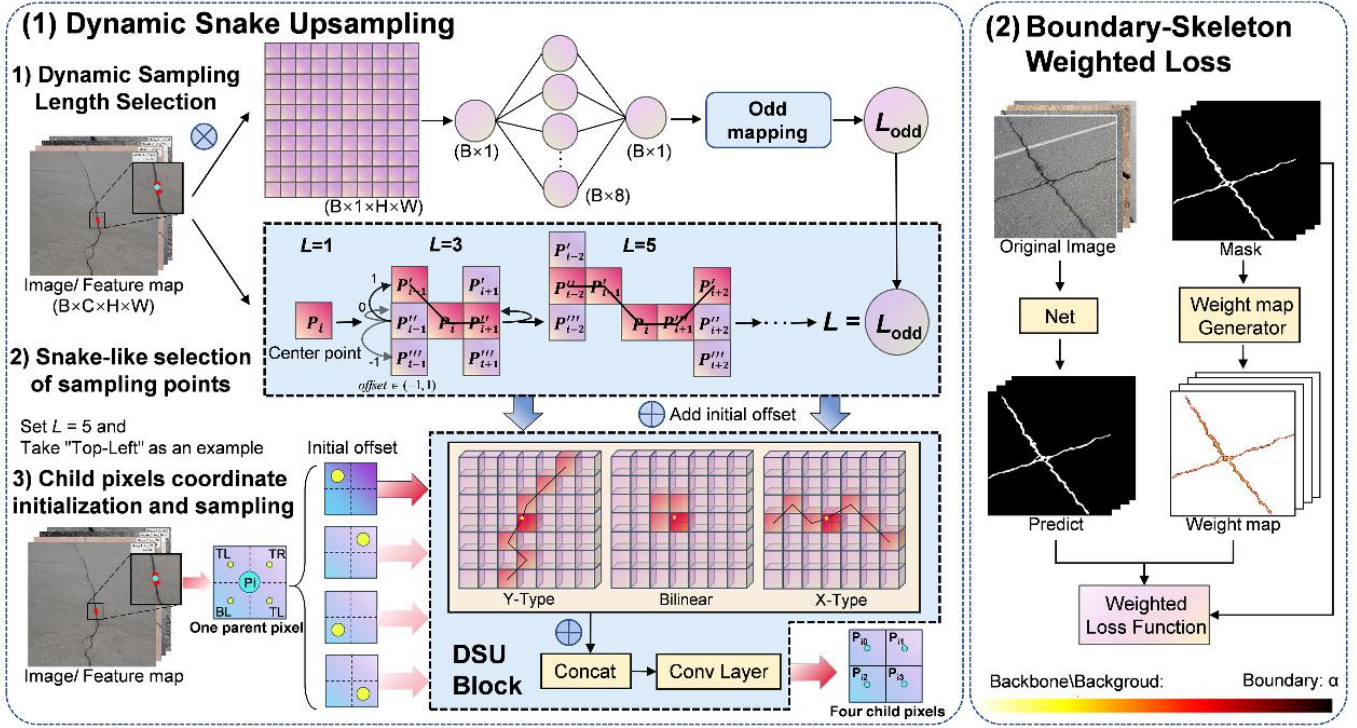

**Fig. 1.** DSU and BSWL methodologies. The left side shows the implementation of DSU with $L = 5$ and selecting the sub-pixel in the upper left corner; the right side shows the implementation of BSWL.

*2.2 Structure-Sensitive Loss*

In image segmentation tasks, class imbalance has long been a critical challenge undermining model performance. For tubular structures specifically, this imbalance compels models to suppress gradient optimization in edge regions to improve overall IoU metrics, frequently resulting in the co-optimization dilemma between regional overlap and topological consistency.

Traditional cross-entropy loss assumes equal importance across all classes, often causing models to disproportionately focus on dominant categories in severely imbalanced tasks. To address this limitation, Dice loss [27] enhances sensitivity to minority-class regions by directly optimizing the Dice coefficient. Tversky loss [38] further improves upon Dice loss for class imbalance through tunable parameters that differentially weight false positives and false negatives. Focal loss [26] dynamically adjusts weights for easily classified samples, effectively mitigating majority-class dominance. While these representative loss functions partially alleviate class imbalance, they still exhibit edge misalignment and structural discontinuity issues.

In recent years, scholars have proposed various boundary- or skeleton-sensitive loss strategies to address these challenges. For boundary-sensitive losses, representative approaches include: Boundary Loss [28], which integrates distance transform maps with prediction outputs to construct a boundary alignment loss function at the differential geometric level, effectively overcoming the sparse gradient issue in boundary regions with traditional segmentation losses. Hausdorff Distance-based Loss [30], which measures geometric discrepancies between boundaries by computing the maximum minimum distance between boundary points. Boundary IoU [39] compensates for the insensitivity of traditional region-based metrics to boundary differences by computing the Intersection over Union within boundary areas.

For skeleton-sensitive losses, notable works include: clDice [29], which introduces a centerline Dice similarity metric computed at the intersection of segmentation masks and skeletons. Persistence Homology-based Methods [31], which extract topological features and design targeted loss functions to improve topological continuity in segmentation results, guiding networks to focus on breakage regions and enhance structural coherence.

Although these methods encourage the model to focus on boundary alignment or topological consistency to varying degrees, the former relies on local gradient sensitivity while the latter depends on global structural reasoning. This leads to a trade-off between boundary ambiguity and structural discontinuity in complex scenarios, often accompanied by a degradation in overall overlap accuracy. DDT [40] partially mitigates this conflict through learnable discrete distance fields for segmentation mask refinement, yet remains constrained by computational complexity.

Inspired by the above work, this paper proposes a precomputed rather than network-generated, continuous rather than discrete distance field weight mapping, which is capable of weighing the model's attention to the boundary versus the subject's structure, and avoids the subject's topological distortions and misalignment of the boundary while suppressing the weakening of the region's overall attention.

### III. METHODOLOGY

*3.1 Dynamic Snake Upsampler*

DSU employs a dual-constraint mechanism: (1) A topology-preserving algorithm based on Dynamic snake convolution [9] that ensures continuity of sampling point sequences through connectivity constraints between adjacent points, and (2) an adaptive scale factor that dynamically optimizes sampling stride via a differentiable adjustment module. We constrain the sampling length to a continuous set of odd numbers, with experimental results in Table 1 demonstrating that {3, 5, 7, 9} constitutes a reasonable set. The DSU implementation is as follows:

**(1) Subsampling Point Coordinate Initialization**

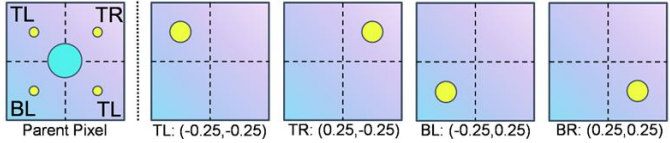

**Fig. 2.** Initial offset of new sampling point coordinates. T for Top, B for Bottom, L for Left and R for Right.

In the dynamic upsampling process, the initial position of the subpixel points is determined by the coordinates of the parent pixel points (x, y) and the preset initial offset  and . As shown in Fig.2, taking 2× upsampling as an example, each original pixel generates four new sampling points, whose initial position offsets can be represented as in Fig.2 above. Here, represent the initial offsets of new sampling points relative to the original pixel in horizontal and vertical directions respectively. This design of initial offsets makes the positions of new sampling points better conform to their initial distribution characteristics, thereby enabling more accurate offset and interpolation based on initial positions during subsequent snake dynamic sampling, and better adaptation to local features of specific input data.

**(2) Dynamic Selection of Sampling Stride**

The sampling stride represents the maximum axial extension distance from the sampling center to the sampling point in the upsampler. Through dynamic adjustment of the sampling stride, the optimal sampling range can be adaptively determined on feature maps of different scales, thereby optimizing feature capture quality. Specifically, an excessively long sampling stride may introduce irrelevant feature information from unrelated regions, leading to enhanced noise and affecting the model's discriminative ability. On the other hand, a sampling stride that is too short restricts the sampler to local neighborhoods, making it difficult to capture broader global information, thus affecting feature fusion. Therefore, based on the adaptive sampling strategy of the feature map, the sampling stride can be dynamically adjusted according to the hierarchical changes in the feature map, helping to balance local information retention and global information fusion at different scales.

The specific implementation is as follows:

**Step1:** The input feature map is compressed into a scalar feature $z$:

$$z = \frac{1}{HW}\sum_{i=1}^{H}\sum_{j=1}^{W}(W_c \cdot X_{:,:,i,j}) \quad (1)$$

**Step2:** The scalar feature is transformed and mapped to an odd value:

$$L_{dy} = L_{base} \cdot (1 + 0.5 \cdot \tanh(W_2^T \cdot \text{ReLU}(W_1 \cdot z))) \quad (2)$$

$$L_d \xrightleftharpoons[STE]{round\ to\ odd} L_{odd} \in S_{odd} \quad (3)$$

Where: $X \in \mathbb{R}^{B \times C \times H \times W}$ is the input feature map; $W_c \in \mathbb{R}^{1 \times C}$ is channel-wise compression convolution; $W_2 \in \mathbb{R}^{1 \times C_m}$ and $W_1 \in \mathbb{R}^{C_m \times 1}$ is a fully connected layer, $C_m$ is layer channel dimensions; $S_{odd}$ is a subset of positive odd integers; $L_d$ is the dynamically selected sampling stride (a floating-point value); $L_{base}$ is the user-defined initial stride; STE preserves gradient flow, ensuring $\partial L_{odd}/\partial L_d = 1$, thereby preventing gradient errors during backpropagation caused by non-differentiable odd-number mapping in forward passes.

The following experiments investigate the impact of fixed versus dynamic sampling lengths on sampler performance, thereby determining the range of set $S_{odd}$. Comparative tests were conducted between dynamic sampling lengths and various fixed-length sampling groups. Using a UNet backbone network and the DeepCrack dataset, the results demonstrate that adaptively assigning sampling strides based on feature maps outperforms fixed sampling lengths, enhancing the performance of the DSU upsampling operator.

**TABLE 1** Impact of different sampling strategies on model performance. DY denotes the dynamic sampling stride strategy, while Fixed-x represents fixed sampling stride strategies, where S indicates the preselected fixed stride length.

| Sampling | Fixed | | | | Dynamic |
|---|---|---|---|---|---|
| Style | S=3 | S=5 | S=7 | S=9 | |
| mIoU(%) | 71.92 | 72.21 | 72.24 | 71.85 | **73.24** |

**(3) Dynamic Snake Selection of Sampling Points**

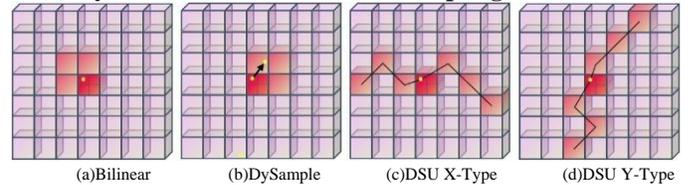

(a)Bilinear  (b)DySample  (c)DSU X-Type  (d)DSU Y-Type

**Fig. 3.** Comparison of Different Upsampling Operators. Sampling regions (red pixels) for target sub-pixels (yellow dots) using different upsampling operators, where black lines represent snake-like sampling paths.

To better align the sampling path with tubular structures, inspired by dynamic snake convolution, we introduce a continuously constrained iterative strategy to sequentially select sampling points. This forces the module to expand the sampling domain, thereby obtaining more suitable information for upsampled sub-pixels. Taking X-shaped sampling as an example, we dynamically determine the sampling stride using the aforementioned stride selection method. Starting from the sub-pixel center, we enforce a unit pixel distance extension along the X-direction while implementing oscillating iterative offsets in the Y-direction, as shown in (4) and Fig.4, where $c = L_{odd} // 2$:

$$P_{i\pm c} = \begin{cases}(x_{i+c}, y_{i+c}) = (x_i + \Delta x_0 + c, y_i + \Delta y_0 + \sum_i^{i+c} \Delta y) \\ (x_{i-c}, y_{i-c}) = (x_i + \Delta x_0 - c, y_i + \Delta y_0 + \sum_{i-c}^{i} \Delta y)\end{cases} \quad (4)$$

the equation for Y-shaped sampling is given in (5):

$$P_{j\pm c} = \begin{cases}(x_{j+c}, y_{j+c}) = (x_j + \Delta x_0 + \sum_j^{j+c} \Delta y, y_j + \Delta y_0 + c) \\ (x_{j-c}, y_{j-c}) = (x_j + \Delta x_0 + \sum_{j-c}^{j} \Delta y, y_j + \Delta y_0 - c)\end{cases} \quad (5)$$

Along the X-shaped or Y-shaped serpentine paths (black lines), upsampling points (endpoints of the black lines) are selected. After reordering the selected sampling points, a convolution layer with a kernel size of (sampling stride, 1) or (1, sampling stride) is applied to obtain the feature value of the new sampling point. Unlike bilinear interpolation (Fig.3(a)) and DySample (Fig.3(b)), this operator no longer rigidly selects four original pixel values in a fixed square neighborhood for linear interpolation. As shown in Fig.3(c) and (d), DSU selects more representative sampling points based on feature similarity to determine the new feature value. Compared with conventional dynamic upsampling (Fig.3(b)), DSU better adapts to fine topological tubular structures by integrating nearby features and assigning more appropriate feature values to new sampling points.

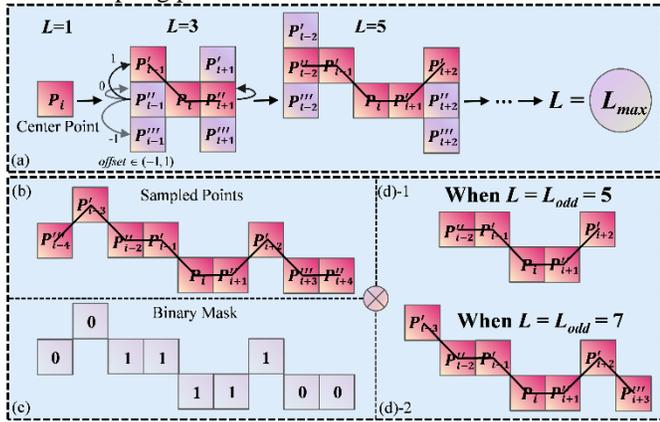

**Fig. 4.** Schematic diagram of DSU's progressive sampling point selection. DSU selects sub-pixel-level sampling points along the *offset* direction, where the *offset* value ranges between (-1, 1). DSU iterates progressively from $L=1$ to $L=L_{ood}$.

In the implementation of the sampling process, we propose a masking mechanism to dynamically select sampling points in order to adapt to varying sampling strides. Since different network layers have different requirements for sampling density, excessively long strides may lead to information redundancy. To address this, we mask out unnecessary positions based on current needs, preserving only feature responses that are relevant to the task..

### 3.2 Boundary-Skeleton Weighted Loss

The traditional Dice loss function treats prediction errors of all pixels equally. However, in topological tubular structure segmentation tasks, mis-segmentation of pixels in either the main body or boundary regions can lead to topological structure breaks, morphological distortion, and boundary misalignment issues. To address this, we proposes BSWL, a dual-constrained Dice loss function that dynamically adjusts pixel weights to enhance the model's focus on important regions. The following describes how to precompute the boundary-skeleton weighting map.

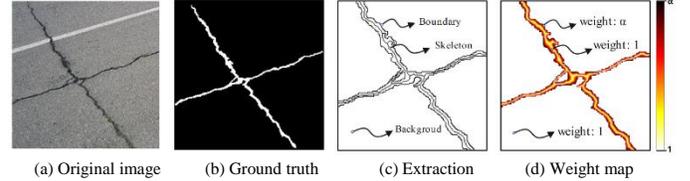

(a) Original image  (b) Ground truth  (c) Extraction  (d) Weight map

**Fig. 5.** Illustration of the weight map generation process. (a) Original image; (b)Ground truth; (c) Extraction: The extracted boundary and skeleton regions; (d) Weight map: The weight map in the form of a heatmap, where the background and boundary are assigned a weight of 1, the skeleton is assigned a weight of $\alpha$, and the weights for the intermediate main regions are linearly interpolated between them.

**Step 1**: Topological tubular structure feature extraction:

The training set mask (b) is processed using morphological operations to extract structural boundaries and skeletons. For boundary extraction, the Canny edge detection algorithm [41] is applied to obtain the outer contour (E) of the tubular structures. For skeleton extraction, the Zhang-Suen algorithm[42] is used to generate a single-pixel-wide medial axis (S), as shown in Fig.5(c).

**Step 2**: Dynamic weight map generation:

$$w(p) = \begin{cases}\alpha - (\alpha - 1) \cdot \dfrac{d_E(p)}{d_S(p) + d_E(p)} & ,\text{if } p \in \Omega_{tube} \\ 1 & ,\text{otherwise}\end{cases} \quad (6)$$

Where: $d_S(p)$ and $d_E(p)$ denote the shortest Euclidean distances from a pixel to the nearest skeleton point and to the outer structural contour, respectively; $\alpha$ is the upper bound of the boundary weight; $\Omega_{tube}$ represents the enclosed continuous space of the tubular structure formed by $\Gamma$, where $\Gamma$ denotes the set of closed contours of the tubular structure obtained through edge detection.

The experiments (Tables 2 and 3) show that when $\alpha$ is approximately equal to the negative-to-positive sample ratio of the dataset, both clDice and Assd demonstrate significant improvements. However, when $\alpha$ is too large, all metrics degrade. To balance "region reconstruction" with "topological structure validity and boundary alignment precision," it is empirically set to approximately match the negative-to-positive sample ratio. The background-to-crack pixel ratio of the DeepCrack dataset is 33.12, and the background-to-vessel pixel ratio of the DRIVE dataset is 9.09. In subsequent experiments on the DeepCrack dataset, $\alpha = 20$, and for the DRIVE dataset experiments, $\alpha = 10$.

**TABLE 2** Performance on the DeepCrack dataset under different values of $\alpha$, region overlap (mIoU), topological continuity (clDice), and boundary alignment accuracy (Assd).

| $\alpha$ | mIoU(%) | clDice(%) | Assd ↓ |
|---|---|---|---|
| 0 | 75.18 | 89.13 | 3.948 |
| 10 | 75.05 | 89.71 | 3.804 |
| **20** | **75.46** | 90.03 | 3.568 |
| 30 | 75.17 | **90.21** | **3.491** |
| 40 | 74.67 | 89.24 | 3.952 |

**TABLE 3** Performance on the DRIVE dataset under different values of $\alpha$, region overlap (mIoU), topological continuity (clDice), and boundary alignment accuracy (Assd).

| $\alpha$ | mIoU(%) | clDice(%) | Assd ↓ |
|---|---|---|---|
| 0 | 67.96 | 83.29 | 0.965 |
| **10** | **68.54** | **83.89** | **0.931** |
| 20 | 68.06 | 83.55 | 0.942 |
| 30 | 68.24 | 83.97 | 0.947 |
| 40 | 68.14 | 83.67 | 0.934 |

**Step 3**: Weighted loss function computation:

Incorporating the weight map into the standard Dice loss function:

$$L_{WeightedDice} = 1 - \frac{2\sum_i w_i p_i g_i + \varepsilon}{\sum_i w_i p_i + \sum_i w_i g_i + \varepsilon} \quad (7)$$

Where $p_i$ denotes the predicted probability, $g_i$ is the ground truth label, $w_i$ is the boundary-skeleton weighting coefficient, $\varepsilon$ is used for numerical stability, introduced to prevent division by zero.

## IV. EXPERIMENTAL SETUP

### 4.1 DataSets

In this paper, we conduct comparative experiments on two representative datasets from different domains: the DeepCrack dataset for road crack detection and the DRIVE dataset for retinal vessel segmentation. During the preprocessing stage, DeepCrack samples are cropped into windowed images with a resolution of 448×448, while DRIVE samples are cropped into 256×256 windows.

### 4.2 Implementation Details

To validate the effectiveness of the proposed DSU and BSWL, both comparative and ablation experiments are conducted. Specifically: (1) To demonstrate the plug-and-play property of DSU and the effectiveness of BSWL, ablation studies are performed using classic segmentation backbones including U-Net, TransUNet, and DSCNet; (2) For performance comparison, different upsampling operators such as bilinear interpolation and dynamic upsampling are integrated into the backbone networks and compared against the proposed DSU module.

For all experiments, the learning rate is scheduled using the polynomial decay strategy [43], with a decay power of 0.9 and a minimum learning rate of 10-6. The batch size is set to 4. To mitigate instability during the early stages of training, a warm-up strategy [44] is applied with 100 warm-up steps. The Adam optimizer is used throughout all experiments.

### 4.3 Evaluation Metrics

To comprehensively evaluate the performance of the segmentation model, this paper uses three complementary metrics, including pixel-level accuracy, structural preservation capability, and boundary alignment precision, as follows:

(1) Mean Intersection over Union (mIoU) [45]: This metric measures the overlap between the predicted segmentation and the ground truth label. It is the most commonly used evaluation metric in semantic segmentation. A higher value indicates more accurate segmentation.

(2) clDice[29]: A structure-aware metric that considers both region overlap and topological consistency of the centerline, particularly suitable for evaluating the segmentation of slender structures such as blood vessels and nerve fibers.

(3) Average Symmetric Surface Distance (Assd)[46]: This calculates the average symmetric distance between the predicted and ground truth boundaries, used to assess boundary alignment accuracy. A smaller value indicates better alignment with the true boundary.

All models are trained on the same dataset, with unified implementation details to ensure fairness.

## V. EXPERIMENTS AND RESULTS ANALYSIS

The ablation experiments demonstrate that with the combined effect of DSU and BSWL, all three classic backbone networks show significant improvements in mIoU, clDice, and Assd metrics, as shown in Table 4. This confirms the universal effectiveness of DSU and BSWL across different datasets and backbone architectures. However, when applying the BSWL strategy to a UNet model using only bilinear interpolation as the upsampling operator, the results were unsatisfactory. While BSWL can balance the model's attention to different structural components, it cannot overcome the inherent information loss caused by the limitations of simple UNet models and bilinear interpolation. Introducing DSU or other advanced upsampling operators effectively addresses this issue.

This paper compares multiple upsampling operators: bilinear interpolation, deconvolution, CARAFE, FADE, SAPA, DySample, and DSU. Experimental results (Table 5) show that DSU achieves the most competitive performance across all metrics (mIoU, clDice, and Assd), particularly demonstrating groundbreaking progress in clDice and Assd. Compared to other upsampling operators, DSU improves clDice by 0.69% to 1.64% and reduces Assd by 0.394 to 0.625.

Under the settings of low-resolution high-channel and high-resolution low-channel input tensors, this paper compares the computational complexity (measured by Floating Point Operations, FLOPs) and the number of parameters (Params) of different upsampling operators. The results show that, unlike other dynamic upsampling operators, there is no significant explosion in computation and parameter count even under high channel numbers or high resolutions (highlighted in red).

To validate that BSWL helps the model to equally focus on "region overlap," "topological continuity," and "boundary alignment," this paper sets up comparison experiments between different loss functions and their combinations with BSWL. Among them, as shown in Table 6. clDiceLoss is a loss function that focuses on topological continuity, and BIoULoss is a loss function that focuses on boundary alignment. While clDiceLoss and BIoULoss perform excellently in improving their respective focus metrics, they weaken the model's ability to learn region overlap. However, the combination of BSWL with these loss functions can achieve collaborative optimization of the three metrics, enabling the model to maintain region overlap (mIoU) while significantly improving topological continuity (clDice) and boundary alignment (Assd).

**TABLE 4  Ablation Experiment. This experiment tests the performance of DSU and BSWL on different backbone networks using the DeepCrack and DRIVE datasets. The values in parentheses next to BSWL in the table correspond to the maximum weight of the weight map.**

| Model | DSU | BSWL | DeepCrack Dataset | | | | DRIVE Dataset | | | |
|---|---|---|---|---|---|---|---|---|---|---|
| | | | mIoU(%) | Dice(%) | clDice(%) | Assd | mIoU(%) | Dice(%) | clDice(%) | Assd |
| DSCNet | | | 75.19 | 85.52 | 89.13 | 3.948 | 67.96 | 80.87 | 83.29 | 0.965 |
| | √ | | 75.69(+0.50) | 85.95(+0.42) | 89.96(+0.83) | 3.307(-0.641) | 68.14(+0.18) | 81.04(+0.17) | 83.62(+0.33) | 0.939(-0.026) |
| | | √ | 75.46(+0.27) | 85.73(+0.21) | 90.04(+0.91) | 3.568(-0.380) | 68.54(+0.58) | 81.29(+0.42) | **83.89**(+0.60) | **0.931**(-0.034) |
| | √ | √ | **75.79**(+0.60) | **86.01**(+0.49) | **90.17**(+1.04) | 3.328(-0.620) | **68.91**(+0.95) | **81.54**(+0.67) | 83.36(+0.07) | 0.943(-0.022) |
| Trans-UNet | | | 72.08 | 83.47 | 88.15 | 4.417 | 67.59 | 80.66 | 82.70 | 1.005 |
| | √ | | 72.53(+0.45) | 83.75(+0.28) | 88.89(+0.74) | 4.300(-0.117) | 67.79(+0.20) | 80.74(+0.08) | 83.39(+0.69) | 0.985(-0.020) |
| | | √ | 73.88(+1.80) | 84.72(+1.25) | 89.00(+0.85) | 4.154(-0.263) | 67.74(+0.15) | 80.76(+0.10) | 83.28(+0.58) | 0.968(-0.037) |
| | √ | √ | **74.57**(+2.49) | **85.26**(+1.79) | **89.26**(+1.11) | **4.031**(-0.386) | **68.36**(+0.77) | **81.14**(+0.48) | **83.78**(+1.08) | **0.955**(-0.050) |
| UNet | | | 71.15 | 82.36 | 86.52 | 4.434 | 62.69 | 77.05 | 78.87 | 1.400 |
| | √ | | 73.42(+2.27) | 83.36(+1.00) | 88.06(+1.54) | 3.836(-0.598) | 68.15(+5.46) | 81.04(+3.99) | 82.97(+4.10) | 0.942(-0.458) |
| | | √ | 72.10(+0.95) | 83.05(+0.69) | 87.36(+0.84) | 4.235(-0.199) | 62.91(+0.22) | 77.18(+0.13) | 78.93(+0.06) | 1.386(-0.014) |
| | √ | √ | **73.76**(+2.61) | **85.49**(+3.13) | **88.94**(+2.42) | **3.692**(-0.742) | **68.41**(+5.72) | **81.23**(+4.18) | **83.46**(+4.59) | **0.935**(-0.465) |

**TABLE 5  Comparison Experiment of Various Upsampling Operators. This includes Bilinear Interpolation, Deconvolution, CARAFE, FADE, SAPA, DySample, and DSU. In the low-resolution high-channel experiment, C=4 and H\W=256, while in the high-resolution low-channel experiment, C=256 and H\W=16.**

| UpSampler | mIoU(%) | clDice(%) | Assd↓ | C=4 , H\W=256 | | C=256 , H\W=16 | |
|---|---|---|---|---|---|---|---|
| | | | | FLOPs(M) | Params(K) | FLOPs(M) | Param(M) |
| Bilinear | 71.15 | 86.52 | 4.434 | 4.19 | 0.000 | 1.05 | 0.0000 |
| TPC | 73.01 | 86.87 | 4.376 | 4.19 | 0.068 | 67.11 | 0.2624 |
| CARAFE | 73.13 | 86.33 | 4.232 | 35.13 | 0.385 | 78.97 | 0.1030 |
| FADE | 72.15 | 86.29 | 4.376 | **4829.22** | 15.006 | 46.03 | 0.0475 |
| SAPA | 73.08 | 86.78 | 4.149 | 90.44 | **656.00** | 22.61 | 0.0339 |
| DySample | 72.10 | 85.92 | 4.463 | 50.33 | 0.640 | **51539.61** | **134.74** |
| **DSU** | **73.36** | **87.56** | **3.838** | 897.58 | 15.356 | 6752.75 | 39.856 |

**TABLE 6  Comparison Experiment of Different Loss Functions and Their Combinations.**

| Loss | mIoU(%) | clDice(%) | Assd↓ |
|---|---|---|---|
| BCE | 75.19 | 89.13 | 3.948 |
| BSWL | 75.46 | 90.04 | 3.568 |
| clDice | 74.38 | 90.55 | 3.404 |
| **clDice+BSWL** | **75.78** | **91.14** | **3.194** |
| BIoU | 75.81 | 90.03 | 3.586 |
| **BIoU+BSWL** | **76.10** | 89.98 | **3.465** |

## VI. VISUALIZATION

To illustrate how the sampling points in the serpentine dynamic upsampling process progressively align with the regions of interest, we extracted the feature maps from the final upsampling stage and visualized the sampling points of the DSU module on the corresponding feature maps. The visualizations cover multiple training epochs (1, 10, 50, 100, 200, and 300). These feature maps demonstrate that, as training progresses, the sampling points gradually shift toward the tubular structures. In the early stages (epochs 1 and 10), the sampling points are relatively scattered and fail to focus effectively on the target regions. As training advances (epochs 50, 100, and 200), the sampling points progressively align with the target structures, exhibiting more precise offset patterns. This confirms DSU's capability to capture topological tubular features and its potential as an effective upsampling operator.

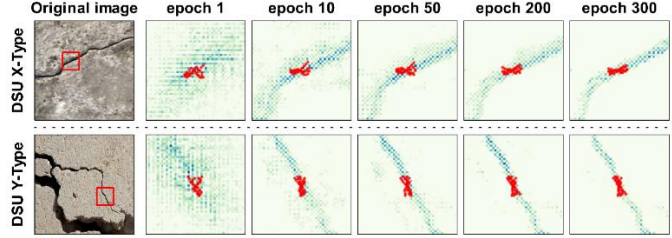

**Fig. 6.** Top-Sampling point offsets of DSU X-Type; Bottom: Sampling point offsets of DSU Y-Type during training. The figure illustrates the distribution changes of the sampling point set (red dots) for a specific point (within the red box in the original image) over the course of training.

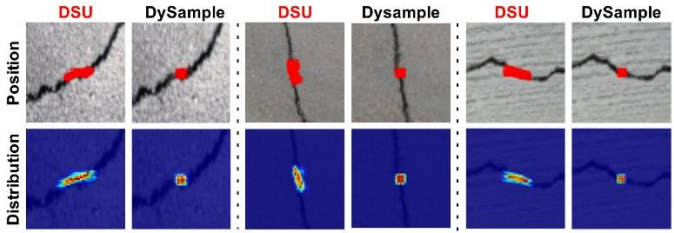

**Fig. 7.** Top-Sampling Point Distribution Map: The coordinates of sampling points are statistically collected and overlaid on the original image to visualize their spatial distribution. Bottom-Sampling Point Attention Heatmap: Based on the learned dynamic offsets from DSU and DySample, we calculate the degree of attention each nearby pixel receives according to sampling bias. The aggregated attention values are then visualized as a heatmap to reflect the focus regions of the upsampling modules.

We selected the most attended point on the feature map along with its 8 neighboring pixels (a total of 9 points) and visualized the coordinates of the sampling points chosen during the upsampling process. Additionally, we generated attention heatmaps to show the sampling point distributions. As illustrated in the figure above, the spatial distribution of sampling points learned by the model aligns more effectively with the structural features of the cracks, especially in high-density regions of the heatmap (i.e., the red areas), where sampling points are biased toward critical features. This allows the model to effectively integrate topological tubular structures and assign meaningful values to target subpixels.

We also visualized the comparative experiments. As shown in Fig.8, the combination of DSU and BSWL enables the model to better preserve structural continuity and avoid topological breakages. Compared with other upsampling operators, DSU demonstrates a superior ability to reconstruct tubular structures with complex topology.

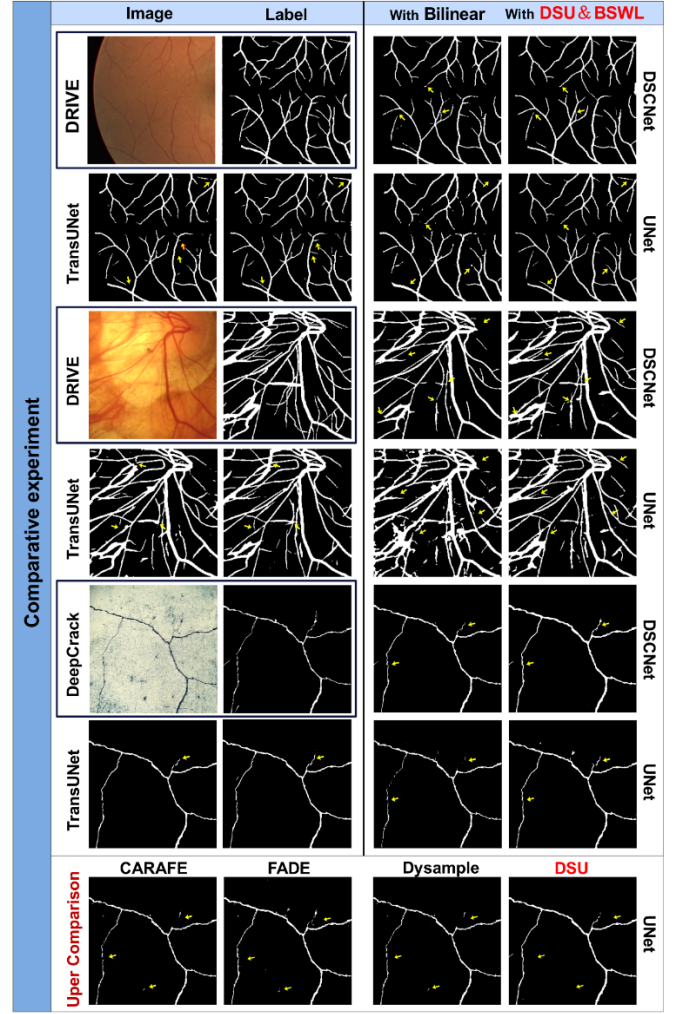

**Fig. 8.** Visualization of Partial Comparative Experiments. Yellow arrows highlight regions where the DSU & BSWL combination demonstrates superior performance

## VII. CONCLUSION

To address the issues of detail loss and topological fragmentation commonly encountered during the upsampling of fine tubular structures, this paper proposes a continuity-constrained and dynamically guided upsampling method, named DSU, along with a precomputed distance field-based loss weighting strategy, named BSWL.

DSU dynamically focuses on the slender branches and boundary regions of tubular structures, adaptively adjusting the sampling range and reconstruction direction. This effectively alleviates the contradiction between fixed sampling kernels and the long-range curvature features of tubular structures. BSWL guides the model in balancing attention between boundaries, skeletons, and the main body regions in between, thus resolving the trade-off between topological completeness and boundary alignment accuracy in segmentation results.

Experimental results show that both methods significantly improve mIoU and clDice while reducing the Assd metric in the segmentation of fine topological tubular structures such as vessels and fissures, indicating substantial improvements in region overlap, topological continuity, and boundary alignment. The synergy between DSU and BSWL optimizes the model's ability in pixel-level prediction as well as topological reconstruction and boundary localization, showing strong potential for applications in fields such as medical image analysis and structural health monitoring.

DATA AVAILABILITY

Data Availability: The data that support the findings of this study are openly available from the corresponding reference papers. For convenience, we have also compiled and organized the relevant data in our GitHub repository, accessible at: https://github.com/YiqiChen02/DSU_dataset.

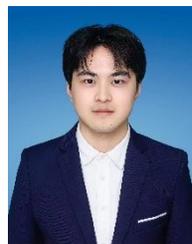

**Yiqi Chen** received his Bachelor's degree in Civil Engineering from Central South University, China, in 2024. He is currently pursuing a Master's degree in Geotechnical Engineering at the School of Civil Engineering, Central South University. His research focuses on deep learning methods for rock mass fracture recognition and digital reconstruction algorithms based on the recognition results. His advisor is Professor Ganghai Huang.

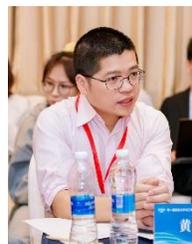

Ganghai Huang received his Master's degree in Mining Engineering from Central South University in 2010 and his Ph.D. in Geotechnical Engineering from the Institute of Rock and Soil Mechanics, Chinese Academy of Sciences, Wuhan, in 2014. He is currently a Professor and Ph.D. advisor at Central South University. He serves as a member of the Committee on Discontinuous Deformation Analysis of the Chinese Society for Rock Mechanics and Engineering. His primary research interests include numerical simulation methods for continuous–discontinuous deformation in multi-field and multiphase engineering materials.

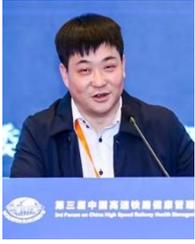
Sheng Zhang received his Ph.D. in Geotechnical Engineering from Nagoya Institute of Technology, Japan, in 2010. He is currently a Professor and Ph.D. advisor at Central South University. He has been awarded the "Excellent Young Scientists Fund" by the National Natural Science Foundation of China and the "Outstanding Youth Fund" by the Natural Science Foundation of Hunan Province. He serves as a member of the Committee on Constitutive Relations and Strength of Soils and the Committee on Transportation Geotechnics of the China Civil Engineering Society, and is a member of the Japanese Geotechnical Society. He is also on the editorial board of several academic journals.

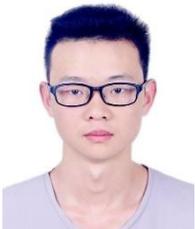
obtained his Bachelor's degree in Civil Engineering from Qinghai University. He is currently a graduate student majoring in Civil and Hydraulic Engineering at the School of Civil Engineering, Central South University. His research direction is the prevention and control of debris flow geological disasters. His advisor is Professor Ganghai Huang.